\pdfoutput=1

\documentclass[10pt, a4paper, twocolumn, showabstract]{naverlabseurope}
\usepackage{times}
\usepackage{latexsym}
\usepackage{lipsum}
\usepackage{multicol}
\usepackage{tikz,tkz-kiviat,pgfplots}
\usepackage{times}
\usepackage{latexsym}
\usepackage{lipsum}
\usepackage{array}
\usepackage{tabularx}
\usepackage[colorinlistoftodos]{todonotes}

\usepackage[T1]{fontenc}

\usepackage[utf8]{inputenc}

\usepackage{times}
\usepackage{latexsym}
\usepackage{lipsum}
\usepackage{array}

\usepackage{subcaption}
\usepackage[colorinlistoftodos]{todonotes}

\usepackage{microtype}
\usepackage{graphicx}
\usepackage{float}
\usepackage{inconsolata}
\usepackage{xcolor}
\usepackage{booktabs}
\usepackage{listings}
\usepackage{multirow}
\usepackage[many]{tcolorbox}
\usepackage{enumitem}

\definecolor{codegreen}{RGB}{93, 168, 128} 
\definecolor{codeblue}{RGB}{74, 74, 250}  
\definecolor{codered}{RGB}{209, 106, 114}
\definecolor{text}{RGB}{68,114,157}

\title{Let your LLM generate a few tokens and you will reduce the need for retrieval}

\correspondingauthor{  herve.dejean@naverlabs.com}

\authors{ Hervé Déjean  }
\affiliations{NAVER LABS Europe}
\contributions{}
\website{https://github.com/naver/bergen}
\websiteref{\href{https://github.com/naver/bergen}}

\tcbuselibrary{breakable,theorems,skins}
\usepackage{hyperref}
\usepackage[many]{tcolorbox}
\usepackage{cleveref}

\newtcbtheorem[no counter,
  crefname={BP}{Best Practices},
  Crefname={BP}{Best Practices}]
{BP}{Recommendation}{%
  enhanced,
  rounded corners,
  colback=text!3!white,
  colframe=text!20!white,
  coltitle=text,
  boxed title style={
    rounded corners,
    size=small,
  } 
}{cor}

\begin{abstract} 
In this paper, we  investigate how efficiently large language models (LLM) can be trained to check whether an answer is already stored in their parametric memory. We distill an LLM-as-a-judge to compute the IK (I Know) score. We found that this method is particularly beneficial in the context of retrieval-assisted augmented generation (RAG), with a respectable accuracy of 80\%. It enables a significant reduction (more than 50\%) in the number of search and reranking steps required for certain data sets. We have also introduced the IK score, which serves as a useful tool for characterising datasets by facilitating the classification task.  Interestingly, through the inclusion of response tokens as input, our results suggest that only about 20,000 training samples are required to achieve good performance. The central element of this work is the use of a  teacher model - the LLM as a judge - to generate training data. We also assess the robustness of the IK classifier by evaluating it with various types of teachers, including both string-based methods and LLMs, with the latter providing better results.
\end{abstract}

\begin{document}
\maketitle
\section{Introduction}

While some limitations of Large Language Models (LLMs) can be alleviated by using Retrieval-Augmented Generation (RAG) \cite{lewis2020retrievalaugmented}, this improvement comes at a computational cost by increasing the prompt length with retrieved documents. An additional drawback of RAG arises when the retrieved information is of poor quality: it degrades the original parametric memory of the LLM \cite{mallen2023trustlanguagemodelsinvestigating} and generates irrelevant answers where a relevant one could be generated without retrieval. Both issues could be avoided by determining whether LLMs need to access this non-parametric additional knowledge to generate their answers. A  solution is to detect when the LLM knows or is confident in its answer. However, it is known that LLMs are generally poor at expressing their faithfulness or uncertainty \cite{xiong2024llmsexpressuncertaintyempirical,Chen2023DoME}.

Relevant work on the \textit{when to retrieve task} can be classified into two main paradigms. On one side, models are trained to recognize (or \textit{probe}) whether an LLM knows the answer or needs to perform the retrieval task. Training material is generated using the LLM itself \cite{labruna2024retrieve,tan_small_2024} or using supposedly strong models such as GPT-4 \cite{Asai2023SelfRAGLT}, which are then distilled into smaller and open models. Models are usually fine-tuned with adapters \cite{labruna2024retrieve}, or by probing some intermediate layers \cite{azaria_internal_2023}.

On the other side, solutions without additional training require paying the cost at inference time with various prompting strategies, such as asking the models to generate multiple answers \cite{zhao_knowing_2024}, or even to generate more elaborated reasoning about these answers \cite{wang_self-dc_2024}. Some metrics are then computed to assess whether the model is confident or not. 

We believe that the context of Retrieval-Augmented Generation (RAG) provides a interesting evaluation framework for research focused on the confidence of LLMs by assessing the effectiveness of these models within the RAG paradigm.

In our approach, we train an LLM to predict its accuracy when answering questions. The training dataset is automatically generated by evaluating the LLM's responses with a 'judge' (either the same LLM or another one). We demonstrate that this setup is sufficient to train an effective 'I Know' (IK) classifier. As in some previous work \cite{tan2024small}, we also allow the model to generate just a few tokens, making the detection task easier. The IK model achieves high effectiveness by correctly identifying when retrieval is needed, thereby improving efficiency by reducing the frequency of retrievals and shortening the average prompt length.


Our contributions are the following:
\begin{itemize}
    \item We distill an LLM-as-a-judge to compute the IK (I Know) score. We have tested and evaluated the distillation with several model families and judges, demonstrating the robustness of this distillation process.
    \item We study the use of  generated answers on the IK score and show that adding 32 or more tokens significantly helps the IK classifier.
    \item We illustrate the impact of the IK score on the final task using several QA datasets. Applying the IK classifier allows for the categorization of datasets based on whether retrieval is needed.
    \item For datasets where retrieval is beneficial, properly setting the IK threshold enables performing retrieval tasks for less than one-third of the queries while improving effectiveness by 2 points.
\end{itemize}

\section{Related Work}
\label{sec:related}
The approaches to the "I Know" task (or its inverse task: "when to retrieve" task) can be divided into two main paradigms. 
\begin{itemize}
    \item One approach involves training models to identify whether the LLM already knows the answer or requires retrieval.
    \item  The other approach requires models to generate multiple answers, followed by the computation of various metrics. Most of these works uses Question Answering datasets for evaluation.
\end{itemize}

The first work training a model for the IK task is the Self-RAG system \cite{Asai2023SelfRAGLT}, where an LLM is trained using GPT-4 as teacher.
\cite{labruna2024retrieve} is an example where the finetuning task combines both the IK task and the Question answering task. 
Instead of finetuning or adapting a LLM, \cite{azaria_internal_2023} uses intermediate layers  outputs as input for a Multi Layer Perceptron. In our experiments this setting underperforms our setting (finetuning the full LLM with an adaptor). 

In \cite{tan_small_2024}, a smaller model (the \textit{judgment model}), hence more efficient, of the same family as the LLM, is trained in order to predict whether this LLM can directly answer the question, assuming  both LLM share the same common knowledge. They also use the generated answer as additional input for their \textit{judgement model}.

Another strategy common to the research on hallucination and confidence estimation detection is to put some more effort on the inference side by using elaborated prompting strategies or several inference steps. 
\cite{ni_when_2024} uses various  prompts with the aim of mitigating overconfidence as well as a Think Step by Step strategy \cite{kojima2022large}.
\cite{jiang2023active} presents FLARE (Looking Active REtrieval augmented generation). FLARE iteratively generates a temporary next sentence, uses it as the query to retrieve relevant documents if it contains low-probability tokens and regenerate the next sentence until reaches the end.
\cite{Ding2024RetrieveOW} proposes the Rowen system which generates multilingual answers and \textit{evaluates the consistency of the perturbed responses across various languages for the same queries.}
In a more elaborate pipeline \cite{tan_small_2024} reformulates the original query into  $n$ rewritten queries and a retrieval is done based on these queries.
The  self-detection system \cite{zhao_knowing_2024} generates several answers and computes their  consistency score corresponding to its confidence.
Self-Eval \cite{kadavath_language_2022} asks for the LLM a self assessment using several generated answers but also evaluates whether models can be trained to predict "P(IK)", the probability that "I know" the answer to a question.
The Self-Ask system \cite{press2022measuring} iteratively prompts the LLM to determine whether to produce follow-up questions for further queries or to directly generate the final answer.
\cite{wang_self-knowledge_2023} employs a retriever to fetch the top-k nearest neighbor questions from the training set, deciding whether retrieval is necessary based on the number of neighboring questions that do or do not require retrieval.


\begin{table*}[htb]
\footnotesize
    \centering
    \caption{Results for the four models are presented, showcasing accuracy and AUC for the IK task with both 0 and 32 tokens from the generated answer. We demonstrate the impact on the QA task by providing the LLMEval score when the IK threshold is set to 0.5 \(IK = 0.5\). Additionally, we identify the \textbf{best point} prducing the best LLMEval, along with the corresponding retrieval percentage and  IK threshold. KILT NQ dataset (dev split). LLMEval(Solar) used as QA metric.}
\begin{tabular}{lc||cc||cccccccc}
\toprule
 \bfseries Model & \bfseries Resp length &\bfseries  ACC & \bfseries AUC & \bfseries NORAG &\bfseries RAG &\multicolumn{2}{c}{\bfseries IK$=$0.5}  &\multicolumn{3}{c}{\bfseries Best Point} \\
    &  & & & & &\bfseries LLMEval&\bfseries \%Retr &\bfseries  LLMEval &\bfseries  \% Retr &\bfseries IK $\theta$ \\              
    \midrule
    
        \midrule
    Meta-Llama-3.1-8B-Instruct  & 0  & 0.74 & 0.81 &0.61 & 0.75& 0.74&40 & 0.76&60&0.7	\\         
    Meta-Llama-3.1-8B-Instruct  & 32  & 0.81 & 0.90 & 0.61 & 0.75 & 0.77&40 & 0.78&55&0.7\\         
    \midrule
    Mistral-7B-Instruct-v0.2  & 0 & 0.77 & 0.84 & 0.62 & 0.76 & 0.76&39 & 0.78&69&0.7 \\         
    Mistral-7B-Instruct-v0.2  & 32  & 0.82&0.89  &0.62 & 0.76 & 0.77&36 & 0.79&50&0.8 \\       
        \midrule
    gemma-2-9b-it  & 0  &0.78 & 0.86  & 0.60 &0.77 & 0.75&40 & 0.77&60&0.7\\         
    gemma-2-9b-it & 32  &0.83 & 0.92  & 0.60& 0.77 &0.77&40 & 0.79&50&0.7    \\         
        \midrule
    SOLAR-10.7B-Instruct-v1.0  & 0  &0.790 & 0.83& 0.69& 0.81 & 0.79&27 & 0.81&60&0.8 \\       
    SOLAR-10.7B-Instruct-v1.0  & 32  & 0.82 & 0.89 &0.69 &0.81 & 0.80&29 & 0.82&48&0.8  \\   
    \midrule
    \bottomrule
    \end{tabular}
    \label{tab:main}
\end{table*}

  \begin{figure*}[]
     \centering
     \includegraphics[width=\linewidth]{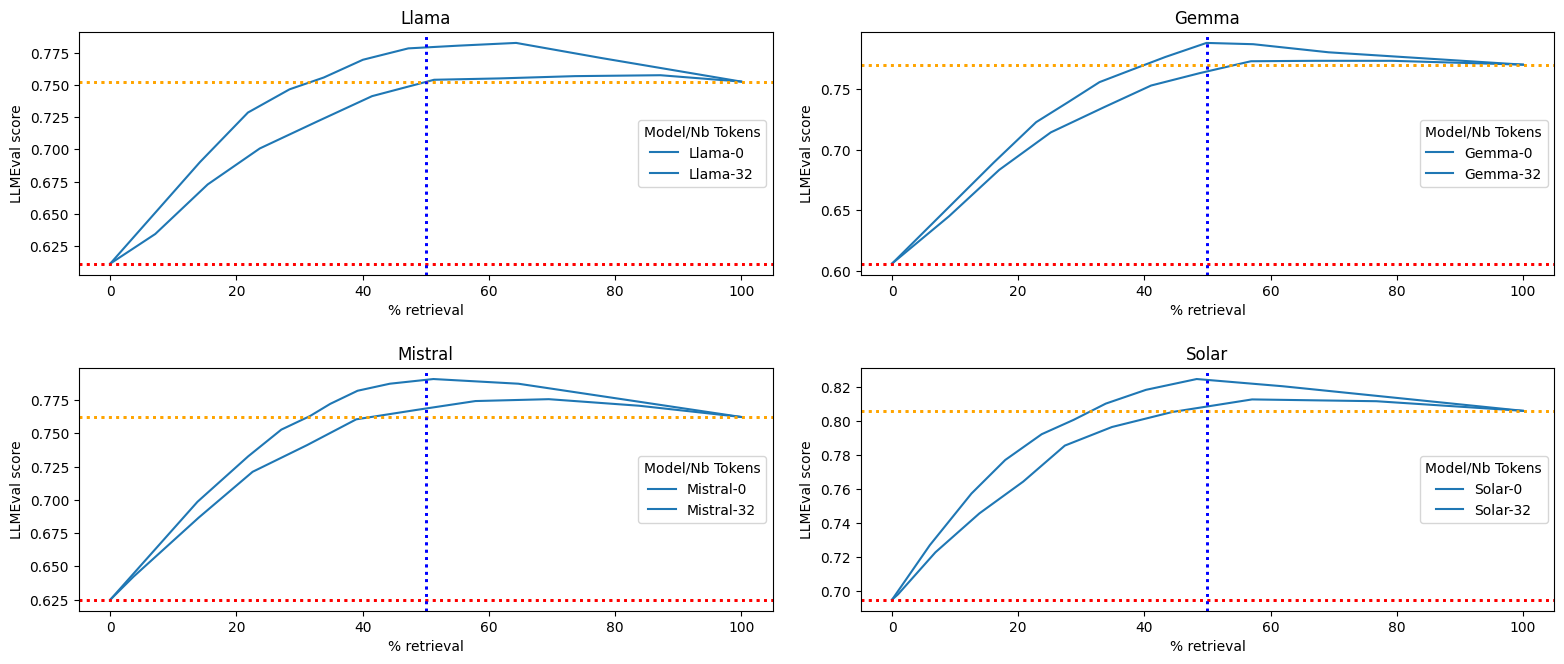}
     \caption{Visualisation of the results shown Table~\ref{tab:main}. The red dotted line indicates the LLMEval score for the No RAG scenario, while the yellow dotted line represents the RAG score. The vertical blue dotted line marks a 50\% retrieval rate. The x-axis shows the percentage of retrieval conducted. Results are displayed for both 0 and 32 tokens from the answer.} 
     \label{fig:4models}
 \end{figure*}

\section{Experimental Methodology}
Our goal is to train a language model to predict whether it can answer a question without needing non-parametric knowledge, such as retrieved documents. We first describe how we build our training set by distilling a large language model (LLM) used as a judge. The task requires the LLM to determine whether it can answer the question without employing retrieval-augmented generation (RAG) by generating a specific token, either 'Yes' or 'No'. The generated "IK" score indicates whether the retrieval mechanism should be activated. 

The final evaluation uses the BERGEN benchmarking library \cite{rau2024bergen} to assess performances across various Question Answering datasets: NQ \cite{seo-etal-2019-real}, ASQA \cite{stelmakh-etal-2022-asqa}, POPQA \cite{mallen2023trustlanguagemodelsinvestigating},  Trivia QA \cite{joshi-etal-2017-triviaqa}, HotpotQA \cite{yang-etal-2018-hotpotqa}, ans SCIQ \cite{welbl-etal-2017-crowdsourcing}.

\subsection{Training Data}

To teach the LLM what it knows (IK score), we apply the model to a dataset and evaluate its output using an LLM-as-a-judge, as described in recent works \cite{yan2024llm-evaluator,Asai2023SelfRAGLT,labruna2024retrieve,tan2024small}. Although this approach is still controversial \cite{Thakur2024JudgingTJ,huang_limitations_2024}, it allows us to achieve satisfactory performance for the IK task and the final task (QA). The score generated by the LLM-as-a-judge is used as a binary label (Yes or No). If the judge returns a numerical score, it is converted into a binary form. To evaluate the importance of the choice of the LLM as a judge, we generated several sets of scores using different LLMs or string-base d metrics (Section~\ref{sec:teacher}).

In the BERGEN framework, the judge uses the question, the generated answer, and the gold answer to provide its judgment, necessitating labeled training data. For our experiments, we use the NQ dataset \cite{seo-etal-2019-real} as the training set. This training set comprises 83,000 questions, 4,000 of which are selected as the validation set. We run the LLM on this set of questions, and the LLM-as-a-judge is then applied to the output to generate the silver labels. In Section~\ref{sec:training}, we investigate the impact of the size of this training set by experimenting with various sizes.

\subsection{Training task}
The training task simply involves asking the LLM to predict the words 'Yes' or 'No' corresponding to the silver labels generated by the judge. We fine-tune the LLMs using adapters for LLM that are 7B or larger (the smaller are fully finetuned), running one epoch with a learning rate of $1e-4$ (specific LR is needed for the smaller models). 

The final IK score is computed as follows: the softmax function is only applied to the logits of the 'Yes' and 'No' tokens, and the score of the 'Yes' token is considered as the IK score (similarly to the LLMEval score in BERGEN). 

We explore two scenarios: one where the input consists solely of the question, and another where it includes the question plus a specified number of tokens generated by the LLM (16, 32, 64, or 128 tokens). The training process takes roughly one hour on a single A100 GPU. Variation between runs is minimal (less than 0.005), so the results presented are based on a single run.

Other configurations, such as those described in \cite{azaria_internal_2023}, were also tested. However, this straightforward and intuitive approach, similarly employed in \cite{Asai2023SelfRAGLT,tan2024small}, has proven to be the most effective.

To examine the behaviors of different LLM families, we selected Meta-Llama-3.1-8B-Instruct, Mistral-7B-Instruct-v0.2, Gemma-2-9b-it, and SOLAR-10.7B-Instruct-v1.0 (using Hugging Face naming conventions). Although SOLAR is based on Mistral, it was selected for being the top-performing model in the BERGEN benchmark, but also because we use it as our main LLM-as-a-Judge. 
For consistency, we use the instruct version and chat template for all models. For the IK task, We compute metrics such as accuracy and the area under the ROC curve (AUC). Evaluations are conducted on the test split for all datasets. 

\subsection{Question Answering Task}

Beyond assessing the pure \textit{I Know} (IK) task, we also conduct an evaluation in the "When To Retrieve" context. In this setting, the IK score is used to decide whether or not to trigger the retrieval-augmented generation (RAG) mechanism, thereby enhancing the prompt with retrieved documents.

We use BERGEN \cite{rau2024bergen}, an end-to-end library for reproducible research that standardizes RAG experiments. With this library, we generate answers in both settings: with and without RAG. 
As  BERGEN settings, we use Splade as the retriever and a DeBERTa-V3-large as the reranker. 50 documents are retrieved and reranked, and the top 4 documents are used in the RAG prompt. Wikipedia serves as the document collection, with pages divided into chunks of 100 tokens each. The generated answers are a maximum length of 128 tokens.

Evaluation is performed using the LLMEval score provided by BERGEN. For each query, the IK score is computed, and depending on a predefined threshold, either the No RAG or the RAG answer is selected. The final evaluation is conducted on the set of selected answers.

\section{Evaluation}
We begin by assessing the IK task across various model families (Section \ref{sec:ik}) and datasets (Section \ref{sec:qa}). 
Additionally, we examine how the number of tokens in the generated answer (Section \ref{sec:length}) affects performance as while as the amount of training material (Section \ref{sec:training}).

\subsection{Main Results}
\label{sec:ik}
Table~\ref{tab:main} shows the global  results for the four models with no answer and considering 32 tokens  of the generated answers. The caption explains how to read the last three columns.
Overall, using the IK score to activate the RAG setting results in similar or improved performance while avoiding the activation of the RAG setting for 50\% of the cases.
When provided with only the question as input, the IK classifiers achieve an accuracy of 75-78\%. Although not exceptional, this level of effectiveness is sufficient to produce good enough results for the QA task, delivering similar outcomes while saving 50\% on RAG usage.
When we include the first 32 tokens of the generated answers, we observe an improvement not only in efficiency but also in effectiveness, with an increase of 2-3 absolute points. 
Figure~\ref{fig:4models} illustrates how the LLMEVal score varies with the percentage of retrieval operations for the four models.

For the NQ dataset, using a threshold of 0.5 proves to be very efficient. However, a threshold of 0.7 offers a great compromise, maintaining efficiency while also enhancing effectiveness.

\subsection{Generalisation to other datasets}
\label{sec:qa}

\begin{figure*}[]
     \centering
     \includegraphics[width=\linewidth]{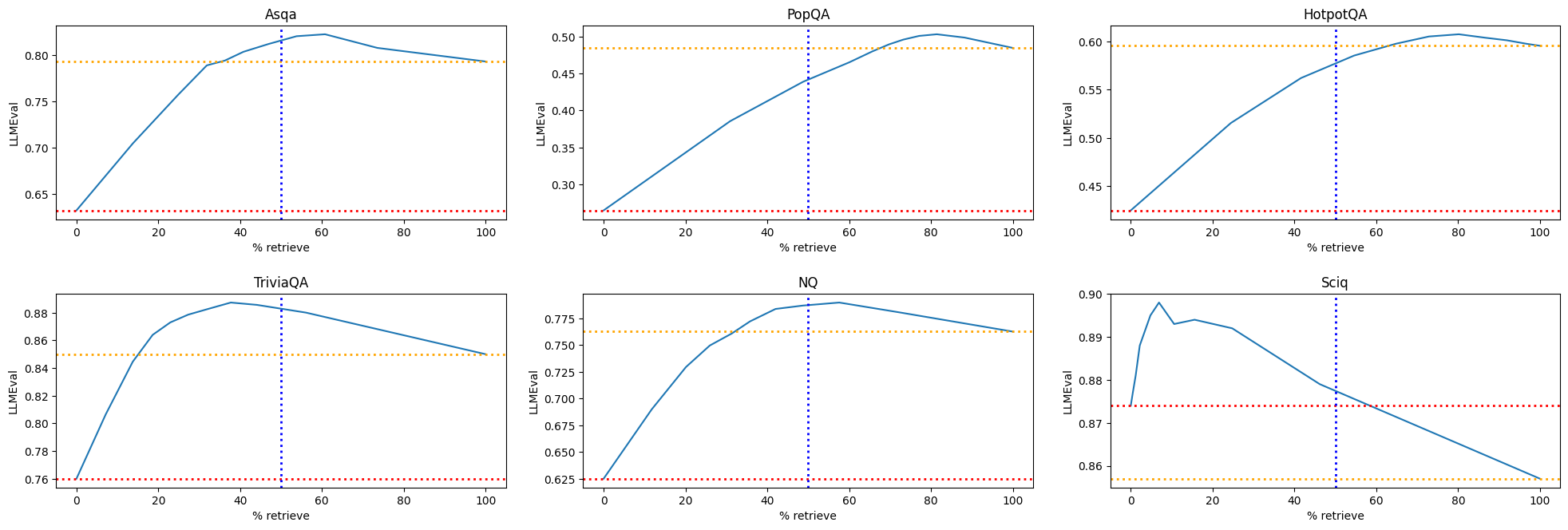}
     \caption{Evaluation conducted on six datasets, with the IK model trained using the NQ dataset. Red dotted line correspond to the NO-RAG result, orange one to the RAG results. See also Table~\ref{tab:datasets}. }
     \label{fig:datasets}
 \end{figure*}

Table~\ref{tab:datasets} shows the application of the IK classifier obtained with the Mistral model  (row 4 of Table~\ref{tab:main}) on  our other datasets. 
Figure \ref{fig:datasets} illustrates the distribution of IK scores across the six selected datasets. Three distinct patterns or behaviors emerge from this visualization.

\begin{itemize}
    \item For both TriviaQA and SCIQ, most IK scores are high, indicating that the LLM is confident in answering questions without resorting to RAG. TriviaQA is specifically designed to evaluate models on general knowledge-based question answering, while the physics, chemistry, and biology questions from the SCIQ dataset also appear to be well within the LLM's knowledge base. This is supported by the high LLMEVAL scores observed in the No RAG (closed-book) setting. Figure~\ref{fig:datasets} shows the strange behaviour of the SCIQ dataset: no RAG is better than RAG. The collection (wikipedia) may be not suited for this dataset(?).
    \item For PopQA and HotpotQA, the IK scores are low, which aligns with the nature of these datasets: PopQA focuses around questions involving infrequent entities or facts, while HotpotQA comprises complex questions that necessitate multi-hop reasoning. 
    \item Lastly, both the NQ and ASQA datasets display a U-shaped pattern, with most questions yielding high confidence from the LLM, but a significant portion exhibiting very low IK scores.
\end{itemize}

  \begin{figure*}[]
     \centering
     \includegraphics[width=\linewidth]{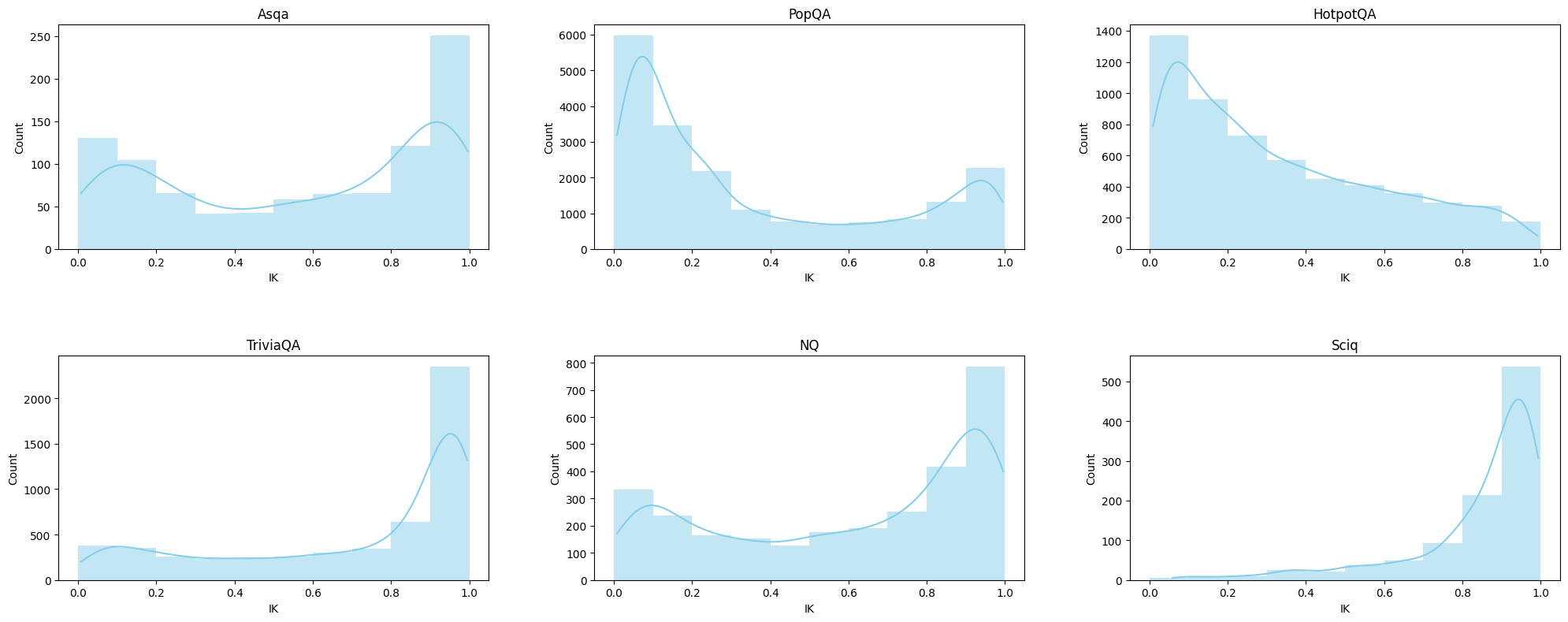}
     \caption{Histogram displaying the distribution of the IK score for six datasets utilizing the Mistral-32 model.} 
     \label{fig:histodatasets}
 \end{figure*}
 
\begin{table*}[htb]
    \centering
    \caption{Evaluation of the Mistral models (Table~\ref{tab:main}, row 3/4), trained on NQ on 5 other QA datasets.}
\begin{tabular}{lccccccccccc}
\toprule
   \bfseries Dataset  & \bfseries  ACC & \bfseries AUC & \bfseries NORAG &\bfseries RAG &\multicolumn{2}{c}{\bfseries IK$=$0.5}  &\multicolumn{3}{c}{\bfseries Best Point} \\
      & & & & &\bfseries LLMEval&\bfseries \%Retr &\bfseries LLMEval &\bfseries \% Retr &\bfseries IK $\theta$ \\    
    \midrule
    NQ &  0.82 & 0.89& 0.62& 0.76 &0.77& 36 & 0.79&50&0.6\\       
    ASQA &0.82 & 0.91 & 0.63& 0.79 & 0.80&35 &0.82& 60 &0.7\\
    HotpotQA &  0.73 & 0.82& 0.42 & 0.60& 0.60& 70& 0.60&70&0.5 \\
    TriviaQA  & 0.86 & 0.92& 0.76 & 0.85 & 0.88&27 & 0.89& 40 &0.8\\
    PopQA& 0.83 & 0.89& 0.27 & 0.48 & 0.49&70 & 0.50& 77&0.8\\
    SCIQ & 0.91 & 0.87& 0.87 & 0.84 & 0.90&7& 0.90& 7&0.5 \\
    \bottomrule
    \end{tabular}
    \label{tab:datasets}
\end{table*}

\subsection{Impact of the Response Length}
\label{sec:length}

As mentioned Section \ref{sec:related} it is often observed that generated answers help improve the input provided to models. In our experiment, the maximum length for an answer is set at 128 tokens. We assess the IK task across a range of answer lengths from 0 to 128 tokens, considering that only part of the generated response might be necessary to determine whether the LLM genuinely knows the answer or is being deceptive. Additionally, we believe that it may be more efficient to halt the LLM as early as possible and then assess whether RAG activation is needed.
Table~\ref{tab:length} shows the impact of the length of the generated answer. If a too small number of tokens (4-16) is not very helpful, considering the full length is also not mandatory. A length of 32 or 64 tokens is helpful enough. 

 \begin{figure}[]
     \centering
     \includegraphics[width=\linewidth]{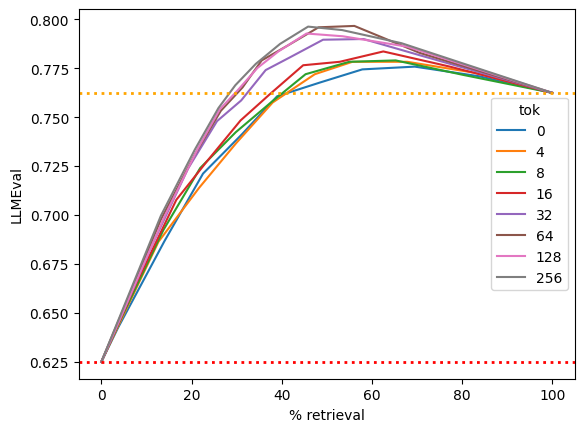}
     \caption{Impact of the response length used during IK task: a minimal number of tokens (32) is required but the plateau is quickly reached: the gain between 32 and 128 is marginal. See Table~\ref{tab:length}.}
     \label{fig:responselength}
 \end{figure}

\begin{table*}[htb]
    \centering
    \caption{Impact of the number of considered response tokens. NQ dataset, Mistral model. While not visible on the LLEval score the increase of used tokens has a positive impact on the retrieval percentage.}
\begin{tabular}{lcccccccccc}
\toprule
     \bfseries  Nb tokens  & \bfseries  ACC & \bfseries AUC & \bfseries NORAG &\bfseries RAG &\multicolumn{2}{c}{\bfseries IK$=$0.5}  &\multicolumn{3}{c}{\bfseries Best Point} \\
      & & & & &\bfseries LLMEval&\bfseries \%Retr &\bfseries LLMEval &\bfseries \% Retr &\bfseries IK $\theta$ \\  
    \midrule
     0 & 0.77 &0.84& 0.42& 0.76 & 0.76&40 & 0.78&70 & 0.7\\     
     4 & 0.77 & 0.85 & 0.62 & 0.76 & 0.75&40 & 0.78&55 & 0.7 \\     
      8  & 0.78 & 0.85 & 0.62 & 0.76 & 0.76&40 & 0.78&50 & 0.7 \\
     16&  0.79 & 0.87 & 0.62 &0.76 & 0.76&37 & 0.78&60 &0.8 \\         
     32 & 0.82 & 0.89 & 0.62& 0.76 &0.77&36 & 0.79&50 &0.6\\
     64 &   0.83 & 0.90& 0.62  &0.76 & 0.78&35 & 0.80&60 &0.6  \\         
     128 & 0.83 & 0.90& 0.62& 0.76 & 0.78&34 & 0.79&45& 0.7 \\      
    \bottomrule
    \end{tabular}

    \label{tab:length}
\end{table*}

\subsection{Impact of the Teacher}
\label{sec:teacher}
The next  point we want to investigate is the importance of the teacher used during training. We use as judge the SOLAR model as default judge in BERGEN.  This model has the advantages of being small compared to other 70B judges, hence fast, and according to \cite{rau2024bergen} its performances are well correlated with OpenAi models. 
Since this model is based on a Mistral model, it may be biased towards Mistral models. To explore this potential bias and evaluate the significance of the teacher selection, we trained the Mistral model using a variety of teachers  The first ones are string-based metrics (recall, Match) as well as two larger models: Llama-3.1-70B-Instruct and Mixtral-8x7B-Instruct-v0.1. 
Table~\ref{tab:teacher} shows that the classifier is not able to learn  string-based metrics: a fairly low accuracy of the IK classifier  (73-75\%), leads to poor QA results with 32 tokens. Adding 128 tokens boosts the IK classifier (3 points) but has little impact of the QA task; at IK=$\theta$, 70\% of retrieval is needed.
With respect to the larger LLMs, the results generally align with those of SOLAR, except for Mixtral-8x7B, which yields disappointing outcomes as a judge despite being from the same model family. 
SOLAR appears to be a more lenient judge, permitting higher scores, a behavior confirmed through qualitative analyses. This leniency is necessary for the NQ dataset, where labels might include lists of synonyms or answers comprising multiple items. It might be beneficial to perform this evaluation using a dataset that features a single label, which could simplify the evaluation process.

\begin{table*}[htb]
    \centering
    \caption{Impact of the selected teacher on the Mistral model (32 tokens). NQ dataset, The \textbf{Score} column refers to the metrics used as teacher (M, recall, or LLMEval). 32 tokens are used except (M, Recall/128)}
\begin{tabular}{lcccccc}
\toprule
  \bfseries Teacher  &\bfseries Acc   &\bfseries AUC  &\bfseries RAG  & \multicolumn{2}{c}{\bfseries IK=0.5}\\
     & & & & \bfseries Score &\bfseries \% Retr \\
    \midrule
    M               &0.75  &0.84 &   0.67& 0.66 &70 \\      
    M (128 tokens) & 0.79& 0.86& 0.67&0.68 & 70 \\  
    Recall          & 0.73 & 0.80 &0.62 &0.60 &60\\         
    Recall (128 token) &0.75 & 0.83& 0.62&0.61& 60 \\
    \midrule
    SOLAR as Judge& 0.82 & 0.89 &0.76 &0.79& 35\\    
    Mixtral-8x7B-Instruct-v0.1 as Judge  &0.78&0.82  &0.80& 0.78& 20\\       
    Llama-3.1-70B-Instruct  as Judge  &0.76  &0.85 &0.61& 0.63& 50\\           idem
    GPT-4o as Judge  &   to do & to do &\\ 
    \bottomrule
    \end{tabular}

    \label{tab:teacher}
\end{table*}

\subsection{Impact of the Training Dataset Size}
\label{sec:training}
\begin{figure}[h]
     \centering
     \includegraphics[width=\linewidth]{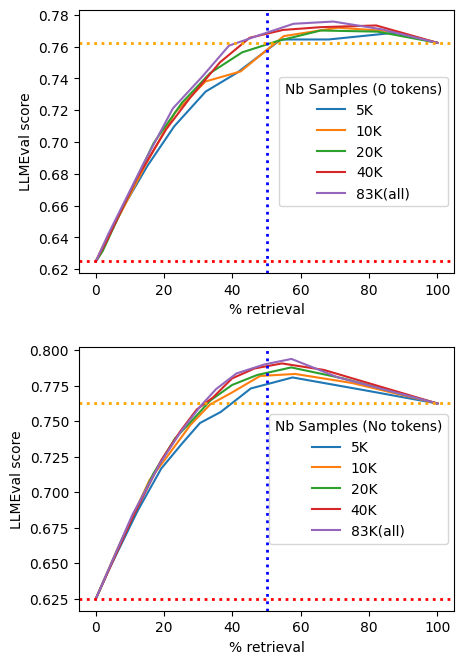}
     \caption{Visualization of the effect of varying the number of samples. A fairly small quantity (5k) already allows for decent performances if 32 tokens from the answer are used. See also Table~\ref{tab:training}. }
     \label{fig:training}
\end{figure}

An interesting question is whether a large training set is necessary to effectively train a model. Table~\ref{tab:training} and Figure~\ref{fig:training} show that the inclusion of tokens in the IK classifier significantly improves performance when the amount of training data is limited. Specifically, with a minimal dataset of just 5K samples, an effective IK classifier can be trained, and with 20K samples, performance levels comparable to those achieved with 83K samples can be attained. In contrast, without these additional tokens, a minimum of 40K samples is needed to reach similar performance levels.

\begin{table*}[htb]
    \centering
    \caption{Reducing the size of the training set leads to decent result, especially when the tokens from the generated answer are used. 
 In this setting (Nb tokens=32), even 5K samples leads to reasonable results. See also Figure~\ref{fig:training}}
\begin{tabular}{lccccccccc}
\toprule
       \bfseries  Nb samples &\bfseries resp. length  & \bfseries  ACC & \bfseries AUC   &\multicolumn{2}{c}{\bfseries IK$=$0.5}  &\multicolumn{3}{c}{\bfseries Best Point} \\
      & (Nb tokens) & &  & LLMEval&\%Retr & LLMEval & \% Retr &IK $\theta$ \\  
    \midrule
    83K &  32& 0.80 & 0.89 &0.77&30& 0.79&50 &0.7\\     
    40K & 32&  0.82 & 0.89 & 0.77&30&0.79&55&0.8\\     
    20K &  32& 0.81 & 0.88&  0.76&30& 0.79&60&0.8\\         
    10K & 32& 0.80 &0.87 & 0.76&30 &0.78&0.7&50 \\    
    5K  & 32 & 0.79 & 0.86& 0.75&30& 0.78&0.8&60 \\
    \midrule
    83k & 0 &0.77 &0.84 & 0.76&40 & 0.78&60&0.7\\
    40k & 0&  0.76 & 0.83 & 0.75&35& 0.77&70&0.7\\
    20k & 0&0.77 & 0.82 & 0.74&30 & 0.77&65&0.8\\
    10k &0& 0.76 & 0.81 & 0.74&30& 0.77&70&0.8\\
    5K &0& 0.74 & 0.80 &  0.70&30 & 0.77&90&0.9\\
    \bottomrule
    \end{tabular}
    \label{tab:training}
\end{table*}

\section{Discussion}
Relying on automatic evaluation with LLM as judge is convenient, but raises doubts about the final evaluation.
How reliable is the LLM-as-judge metrics? Are we not just learning how to increase the judge score instead of detecting a lack of knowledge of the LLM?  A qualitative analysis of the results shows a real improvement in the QA results: the gain is mostly due to cases where non relevant documents were retrieved. This analysis also shows that improving the IK classifier accuracy  would be very beneficial to the QA results, the current level still producing  too many errors.

A detailed study on efficiency is still missing. The latency figures provided should be viewed with caution, as they heavily depend on industrial setups. On average, as seen in Table~\ref{tab:latency}, processing a query without using RAG takes around 18 ms, while incorporating RAG with 5 documents extends this to 78 ms. This implies that skipping the RAG mechanism can potentially reduce processing time by up to 80\%. The IK classifier introduces a minor latency of 3.7 ms per query. In a less-than-optimal scenario, running the large language model (LLM) to generate 32 tokens results in latencies of 30.2 ms (3.7 + 8.3 + 18.2) for the NORAG setup and 90.4 ms (3.7 + 8.3 + 78.4) for the RAG setup. Even with these conditions, the IK option shows a 25\% efficiency gain, when focusing purely on the generation aspect. Furthermore, considering the fastest retriever, BM25, has a latency of 5 ms per query \cite{Lassance_2024}, and the Cohere R3 reranker\footnote{https://cohere.com/blog/rerank-3}  requires 60 ms for reranking 50 documents of 256 tokens, it becomes evident that eliminating the reranking step is vital, as its latency is comparable to that of the generation component in the RAG processing.

\begin{table*}
      \centering
    \caption{Processing time of the KILT NQ dev set (2,837 questions) on a A100 using VLLM (batch size 256). The result with RAG does not consider the retrieval and re-ranking time.}
    \begin{tabular}{lcc}
    \toprule
     \bfseries Nb Generated tokens& \bfseries  total processing time & \bfseries processing time per question (ms)\\
     \midrule
     1  (IK score)& 00:10.56& 3.7 \\
     32 (IK input) & 00:23.58 & 8.3 \\
     128 (NO RAG) &00:51.70 &18.2 \\
     128 (RAG 5 docs) & 03:42.56 &78.4 \\
    \end{tabular}
    \label{tab:latency}
\end{table*}


\section{Conclusion}
We have demonstrated that training a large language model (LLM) to determine whether an answer is stored in its parametric memory is both feasible and beneficial. Even with an 80\% accuracy rate, this method proves valuable in the retrieval-augmented generation (RAG) context, allowing for the majority of retrieval steps to be bypassed for certain datasets. Additionally, the IK score can be leveraged to characterize a dataset. The classification task becomes easier when tokens from the answer are included as input. In this scenario, only a modest amount of training data (20K samples) is necessary. A crucial factor is utilizing an effective teacher model (in this case, the LLM-as-judge) to generate the training data. Improving the quality of the training data would likely enhance the performance of the classifier.

\bibliography{sample}
\bibliographystyle{acl_natbib}
\newpage
\onecolumn
\end{document}